\documentclass{article} 
\usepackage{iclr2017_conference,times}
\usepackage[hidelinks]{hyperref}
\usepackage{url}

\usepackage{natbib}
\usepackage{graphicx}
\usepackage{floatrow}
\usepackage{appendix}

\title{Incorporating long-range consistency in CNN-based texture generation}

\author{Guillaume Berger \& Roland Memisevic \\
	Department of Computer Science and Operations Research \\
	University of Montreal\\
	\texttt{guillaume.berger@umontreal.ca}, \texttt{memisevr@iro.umontreal.ca} \\
}

\begin{document}
	
	\maketitle
	
	\vspace{-0.4cm}
	
	\begin{abstract}
		\citet{Gatys2015b} showed that pair-wise products of features in a convolutional network are a very effective representation of image textures. We propose a simple modification to that representation which makes it possible to incorporate long-range structure into image generation, and to render images that satisfy various symmetry constraints. We show how this can greatly improve rendering of regular textures and of images that contain other kinds of symmetric structure. We also present applications to inpainting and season transfer. 
	\end{abstract}
	
	\vspace{-0.3cm}
	
	\section{Introduction}
	
	There are currently two dominant approaches to texture synthesis:  
    non-parametric techniques, which synthesize a texture by extracting pixels (or patches) from 
    a reference image that are resampled for rendering \citep{Efros02, kwatra}, 
    and parametric statistical models, which optimize reconstructions to match certain statistics computed 
    on filter responses \citep{HeegerB95, Portilla}. 
    Recently, the second approach has seen a significant advancement, after \cite{Gatys2015b} showed that 
    a CNN pre-trained on an object classification task, such as ImageNet \citep{ILSVRC15},
	can be very effective at generating textures. 
    \cite{Gatys2015b} propose to minimize with respect to the input image a loss function, that measures 
    how well certain high-level features of a reference image are preserved.
	The reference image constitutes an example of the texture to be generated.
	The high-level features to be preserved are pair-wise products of feature responses,
	averaged over the whole image, referred to as the ``Gramian'' in that work.
	In \cite{Gatys2015c}, the same authors show that by adding 
	a second term to the cost, which matches the content of another image, one can render
	that other image in the ``style'' (texture) of the first.
	Numerous follow-up works have since then analysed and extended this approach \citep{ulyanov16texture, JohnsonAL16, Ustyuzhaninov2016}.
	
\begin{figure}[t]
		\centering
		\includegraphics[height=2.2cm]{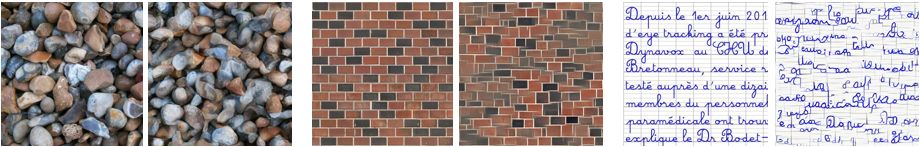}
		\caption{Reference image (\textit{left}) and generated texture (\textit{right}) using the procedure described in \cite{Gatys2015b}.}
		\label{gatys_demo}
	\end{figure}

	As shown in Figure~\ref{gatys_demo}, this method produces impressive results. However, it fails to take into account non-local structure, 
	and consequently cannot generate results that exhibit
	long-range correlations in images.
	An example of the importance of long-range structure is the regular brick wall texture in the middle of the figure. Another example is the task of inpainting, where the goal is to fill in a missing part of an image, such that it is faithful to the non-missing pixels. 
	Our main contribution is to introduce a way to deal with long-range structure
	using a simple modification to the product-based texture features.
	Our approach is based on imposing a ``Markov-structure'' on high-level features,  
	allowing us to establish feature constraints that range across sites instead of 
	being local. 
	Unlike classical approaches to preserving spatial structure in image generation, such as Markov Random Fields and learning-based extensions \citep{Roth2005}, our approach does not impose
	any explicit local
	constraints on pixels themselves. Rather, inspired by \cite{Gatys2015b}, 
	it encourages consistency to be satisfied on high-level features and on average 
	across the whole image. 
	We present applications to texture generation, inpainting and season transfer.
	
	\section{The artistic style algorithm}
	\label{Section1}

	\subsection{Synthesis procedure}
	
	Given a reference texture, $x$, the algorithm described in \cite{Gatys2015b} permits to synthesize by optimization a new texture $\hat{x}$ similar to $x$. To achieve this, the algorithm exploits an ImageNet pre-trained model to define metrics suitable for describing textures: 
	``Gram'' matrices of feature maps, computed on top of $L$ selected layers. Formally, let $N^l$ be the number of maps in layer $l$ of a pre-trained CNN. The corresponding Gram matrix $G^{l}$ is a $N^l \times N^l$ matrix defined as:
	\begin{equation}
	G^{l}_{ij} = \frac{1}{M^l}\sum\limits_{k=1}^{M^l}F^l_{ik}F^l_{jk} = \frac{1}{M^l} \langle\,F^l_{i:},F^l_{j:}\rangle
	\label{eq1}
	\end{equation}
	where $F^l_{i:}$ is the $i^{th}$ vectorized feature map of layer $l$, $M^l$ is the number of elements in each map of this layer, and where $\langle\,\cdot,\cdot\rangle$ denotes the inner product. Equation \ref{eq1} makes it clear that $G^l$ captures how feature maps from layer $l$ are correlated to each other. Diagonal terms, $G^{l}_{ii}$ are the squared Frobenius norm of the $i^{th}$ map $\left\Vert F^l_{i:} \right\Vert_F^2$, so they represent its spatially averaged energy. We will discuss the Gramians in more detail in the next paragraph.
	Once the Gram matrices $\{G^l\}_{l\in[1,L]}$ of the reference texture are computed, the synthesis procedure by \cite{Gatys2015b} 
	amounts to constructing an image that produces 
	Gram matrices $\{\hat{G}^l\}_{l\in[1,L]}$ that match the ones of the reference texture. More precisely, 
	the following loss function is minimized with respect to the image being constructed: 
	\begin{equation}
	\mathcal{L}_{style} = \sum\limits_{l=1}^{L}w_l \left\Vert \hat{G}^l - G^l \right\Vert_F^2 = \sum\limits_{l=1}^{L}w_l\mathcal{L}^l_{style}
	\label{eq2}
	\end{equation}
	where $w_l$ is a normalizing constant similar to \cite{Gatys2015b}. 
	
	\begin{figure}[t]
		\begin{center}
			\includegraphics[width=13cm]{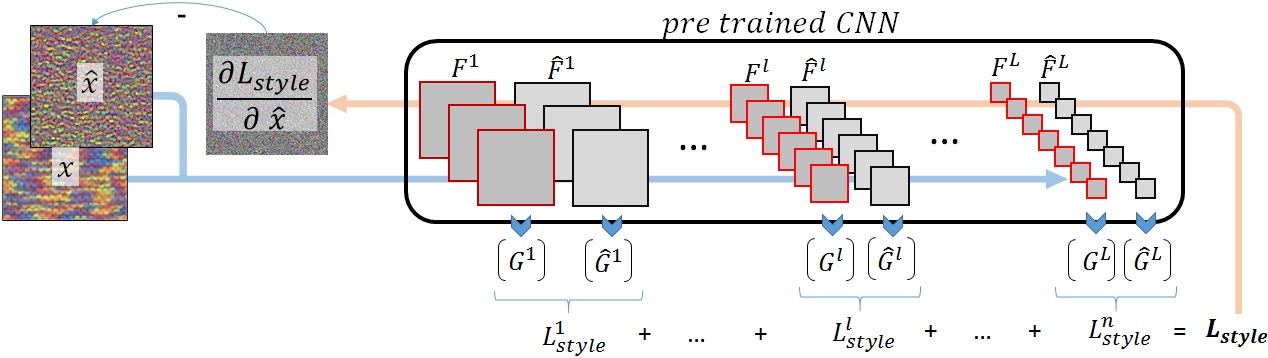}
		\end{center}
		\caption{Summary of the texture synthesis procedure described in \cite{Gatys2015b}. We use a VGG-19 network \cite{SimonyanZ14a} as the pre-trained CNN.}
		\label{synthesis}
	\end{figure}
	
	The overall process is summarized in Figure~\ref{synthesis}. 
	While the procedure can be computationally expensive, there have been successful attempts reported recently which reduce the generation time \citep{ulyanov16texture, JohnsonAL16}.
	
	
	\subsection{Why Gram matrices work}
	
\begin{figure}
		\begin{center}
			\includegraphics[width=13.8cm]{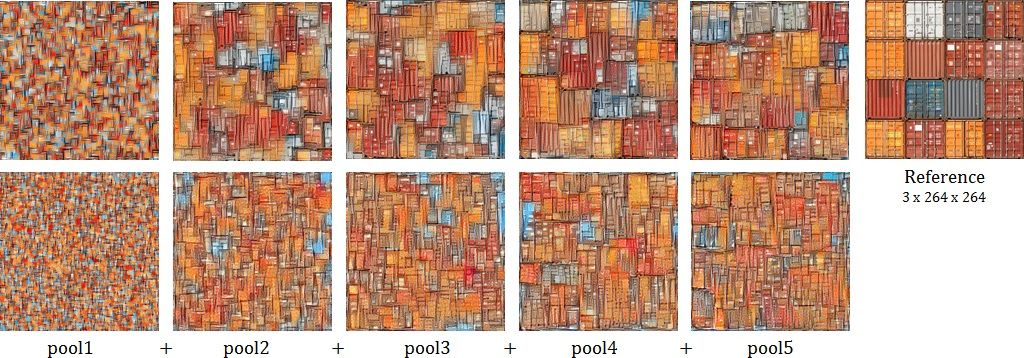}
		\end{center}
		\caption{Exploiting Gram matrices of feature maps as in \cite{Gatys2015b} ($1^{st}$ \textit{row}) or only the squared Frobenius norm of feature maps ($2^{nd}$ \textit{row}) for increasingly deep layers (from \textit{left} to \textit{right}).}
		\label{diag_vs_non_diag}
	\end{figure}

	Feature Gram matrices are effective at representing texture, because they capture 
	global statistics across the image due to spatial averaging. Since textures are static, averaging over positions is required and makes Gram matrices fully blind to the global arrangement of objects inside the reference image. This property permits to generate very diverse textures by just changing the starting point of the optimization. 
	Despite averaging over positions, coherence across multiple features needs to be 
	preserved (locally) to model visually sensible textures. 
	This requirement is taken care of by the off-diagonal terms in the Gram matrix, 
	which capture the co-occurence of different features at a single spatial location. Indeed, Figure~\ref{diag_vs_non_diag} shows that restricting the texture representation to the squared Frobenius norm of feature maps (i.e. diagonal terms) makes distinct object-parts from the reference texture encroach on each other in the reconstruction,  as local coherence is not captured by the model. Exploiting off-diagonal terms improves the quality of the reconstruction as consistency across feature maps is enforced (on average across the image). 
	
	The importance of local coherence can be intuitively understood in the case of 
	linear features (or in the lowest layer of a convolutional network): when 
	decomposing an image using Gabor-like features, local structure can be expressed 
	as the relative offsets in the Fourier phase angles between multiple different 
	filter responses. 
	A sharp step-edge, for example, requires the phases of local Fourier components at 
	different frequencies to align in a different way than a blurry edge or 
	a ridge \citep{Morrone221, Kovesi99}. 
	Also, natural images exhibit 
	very specific phase-relationships across frequency components in general, and destroying
	these makes the image look unnatural \citep{WangS03a}. The same is not true of Fourier amplitudes
	(represented on the diagonals of the Gramian),
	which play a much less important role in the visual appearance \citep{Oppenheim1981}.
	In the case of deeper representations, the situation is more complex, but it is still 
	local co-occurrence \emph{averaged over the whole image} that captures texture. 
	
	Unfortunately, average local coherence falls short of capturing long-range structure in images. Spatial consistency is hard to capture 
	within a single filter bank, because of combinatorial effects. Indeed, since Gram matrices captures coherence at a single spatial location, every feature would have to be matched to multiple transformed versions of itself. A corollary is that every feature would have to appear in the form of multiple transformed copies of itself in order to capture spatial consistency. However, this requirement clashes with the limited
	number of features available in each CNN layer. 
	One way to address this is to use higher-layer features, whose receptive fields are larger. Unfortunately, as illustrated in Figure~\ref{diag_vs_non_diag}, even if using layers up to $pool5$ whose input receptive field covers the whole image\footnote{For this experiment, the image size is $264 \times 264$, which is also the size of the $pool5$ receptive field.} (first row, last column), the reconstruction remains mainly unstructured and the method fails to produce spatial regularities.
	
	
	
	
	
	\section{Modeling spatial co-occurences}
	\label{Section2}
	
\begin{figure}
		\begin{center}
			\includegraphics[height=3.5cm]{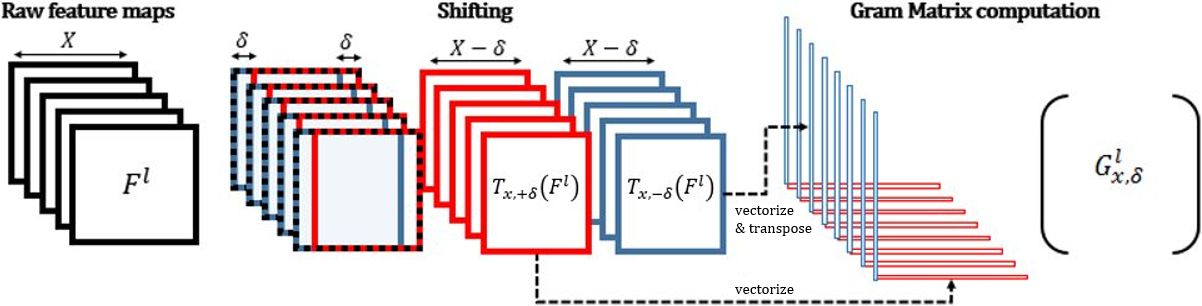}
		\end{center}
		\caption{Computing the shifted Gram matrix for a given layer with feature maps of width $X$.}
		\label{cross_correl_schema}
	\end{figure}

	To account for spatial structure in images, we propose encoding this structure 
	in the feature self-similarity matrices themselves. To this end, we suggest that, instead 
	of computing co-occurences between multiple features within a map, we compute 
	co-occurences between feature maps $F^l$ and 
	\emph{spatially transformed} feature maps $T(F^l)$, where $T$ 
	denotes a spatial transformation. 
	In the simplest case, $T$ represents local translation, which amounts to 
	measuring similarities between local features and other neighbouring features.
	We denote by $T_{x,+\delta}$ the operation consisting in horizontally translating feature maps by $\delta$ pixels and define the transformed Gramian: 
	\begin{equation}
	G^{l}_{x,\delta,ij} = \frac{1}{M^l} \langle\,T_{x,+\delta}\left(F^l_{i:}\right),T_{x,-\delta}\left( F^l_{j:}\right)\rangle
	\label{eq3}
	\end{equation}
	where $T_{x,-\delta}$ performs a translation in the opposite direction. As illustrated in Figure~\ref{cross_correl_schema}, the transformation in practice 
	simply amounts to removing the $\delta$ first or last columns from the raw feature maps.  
	Therefore, the inner product now captures how features at position $(i,j)$ are correlated with features located at position $(i,j+\delta)$ in average. While Figure~\ref{cross_correl_schema} illustrates the case where feature maps are horizontally shifted, one would typically use translations along both the $x$-axis and the $y$-axis. Our transformed Gramians are related to Gray-Level Co-occurrence Matrices (GLCM) \citep{Haralick} which compute the unnormalized frequencies of pixel values for a given offset in an image. While GLCMs have been mainly used for analysis, some work tried to use these features for texture synthesis \citep{Lohmann}. Usually, GLCMs are defined along 4 directions: $0^{\circ}$ and $90^{\circ}$ (i.e. horizontal and vertical offsets), as well as $45^{\circ}$ and $135^{\circ}$ (i.e. diagonal offsets). In comparison, our method does not consider diagonal offsets and captures spatial coherence on high-level features, making use of a pre-trained CNN, instead of working directly in the pixel domain. 
	
	With this definition for transformed Gram matrices, we propose defining the loss as: 
	$\mathcal{L} = \mathcal{L}_{style} + \mathcal{L}_{cc}$, 
	where \textit{cc} stands for \textit{cross-correlation}. 
	Like $\mathcal{L}_{style}$, $\mathcal{L}_{cc}$ is a weighted sum of multiple losses $\mathcal{L}^l_{cc, \delta}$ defined for several selected layers as the mean squared error between transformed Gram matrices of the reference texture and the one being constructed: 
	
	\begin{equation}
	\mathcal{L}^l_{cc, \delta} = \frac{1}{2}\left(\mathcal{L}^l_{cc,x,\delta}+\mathcal{L}^l_{cc,y,\delta}\right)= \frac{1}{2}\left(\left\Vert \hat{G}_{x,\delta}^l - G_{x,\delta}^l \right\Vert_F^2+\left\Vert \hat{G}_{y,\delta}^l - G_{y,\delta}^l \right\Vert_F^2\right)
	\end{equation}

	Although this amounts to adding more terms to a representation that was already high-dimensional and overparametrized, we found that these additional terms do not hurt the diversity of generated textures, even for ones that do not exhibit any spatial regularity. 
	Indeed, the new loss remains blind to the global arrangement of objects. 
	While we focus on translation for most of our results, we shall discuss other types of transformation in the experiments Section.

	\section{Experiments}
	\label{experiments}
	In our experiments, we exploit the same normalized version of the VGG-19 network\footnote{available at http://bethgelab.org/media/uploads/deeptextures/vgg\_normalised.caffemodel.} \citep{SimonyanZ14a} as in \cite{Gatys2015b, Gatys2015c} and layers \textit{conv1\_1}, \textit{pool1}, \textit{pool2}, \textit{pool3},  and \textit{pool4} are always used to define the standard Gram matrices. 
	In our method, we did not use \textit{conv1\_1} to define cross-correlation terms, as the large number of neurons 
	at this stage makes the computation of Gram matrices costly. 
	Corresponding $\delta$ values for each layer are discussed in the next paragraph.
	
	Finally, our implementation\footnote{available at https://github.com/guillaumebrg/texture\_generation} uses Lasagne \citep{lasagne}. Each image is of size $384\times384$. Most textures used as references in this paper were taken from \textit{textures.com} and \textit{pixabay.com}.

	\subsection{Experiments with translation-Gramians}

	The $\delta$ parameter is of central importance as it dictates the range of the spatial constraints. 
	We observed that the optimal value depends on both the considered layer in the pre-trained CNN and the reference texture, making it difficult to choose a value automatically. 
	
	For instance, Figure 5 shows generated images from the brick wall texture using only the \textit{pool2} layer with different $\delta$ configurations. 
	The first row depicts the results when considering single values of $\delta$ only. 
	While $\delta=4$ or $\delta=8$ are good choices, considering extreme long-range correlations does not help for this particular texture: a brick depends mostly on its neighbouring bricks and not the far-away ones. 
	More precisely, a translation of more than $16$ pixels in the \textit{pool2} layer 
	makes the input receptive field move more than $64$ pixels.
	Therefore $\delta=16$ or $\delta=32$ do not capture any information about neighbouring bricks. Unfortunately, this is not true for all textures, and $\delta=16$ or $\delta=32$ might be good choices for another image that exhibits longer structures. 
	
\begin{figure}
		\begin{center}
			\includegraphics[height=5cm]{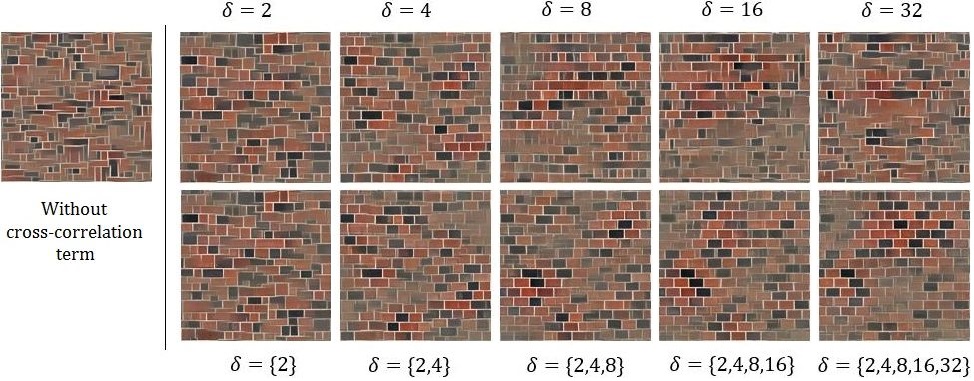}
		\end{center}
		\caption{$Pool2$ reconstruction using different values for $\delta$ in $\mathcal{L}^{pool2}_{cc, \delta}$. (\textit{$1^{st}$ column}): Without cross-correlation terms. (\textit{Other columns}): Using a single cross-correlation term with a fixed $\delta$ value (\textit{$1^{st}$ row}) or using multiple cross-correlation terms with distinct $\delta$ values (\textit{$2^{nd}$ row}).}
		\label{delta_experiment}
	\end{figure}

	Searching systematically for a $\delta$ configuration that works well with the reference texture being considered would be a tedious task: even for a very regular texture with a periodic horizontal (or vertical) pattern, it is hard to guess the optimal $\delta$ values for each layer (for deeper ones in particular). Instead, we propose to use a fixed but wide set of $\delta$ values per layer, by defining the cost to be: $\mathcal{L}^l_{cc} = \sum\limits_{k}\mathcal{L}^l_{cc, \delta_{k}}$. A potential concen is that combining many loss terms can hurt the reconstruction or the diversity of generated textures. Figure~\ref{delta_experiment} (second row) shows contrarily that there is no visual effect from using $\delta$ values that are \emph{not} specifically useful for the reference texture being considered: 
the rendering benefits from using $\delta \in \{2,4,8\}$ while considering bigger values ($\delta \in \{2,4,8,16,32\}$, e.g.) does not help, but does not hurt the reconstruction either. We found the same to be true for other textures as well, and our results (shown in the next Section) show that combining the loss terms is able to generate very diverse textures. A drawback from using multiple cross-correlation terms per layers, however, is computational. We found that in our experimental setups, adding cross-correlation terms increases the generation time by roughly $80\%$.
	
	As a guide-line, for image sizes of roughly $384\times384$ pixels,  
	we recommend the following $\delta$ values per layer (which we used in all our 
	following experiments): 
	$\{2,4,8,16,32,64\}$ for \textit{pool1}, $\{2,4,8,16,32\}$ for \textit{pool2}, $\{2,4,8,16\}$ for \textit{pool3}, and $\{2,4,8\}$ for \textit{pool4}. 
	The number and the range of $\delta$ values decrease with depth because feature maps are 
	getting smaller due to $2\times2$ pooling layers. This configuration should be sufficient to account for spatial structure in any $384\times384$ image.
	
	\subsection{Synthesis of structured texture examples}
	
\begin{figure}[t!]
		\begin{center}
			\includegraphics[width=12cm]{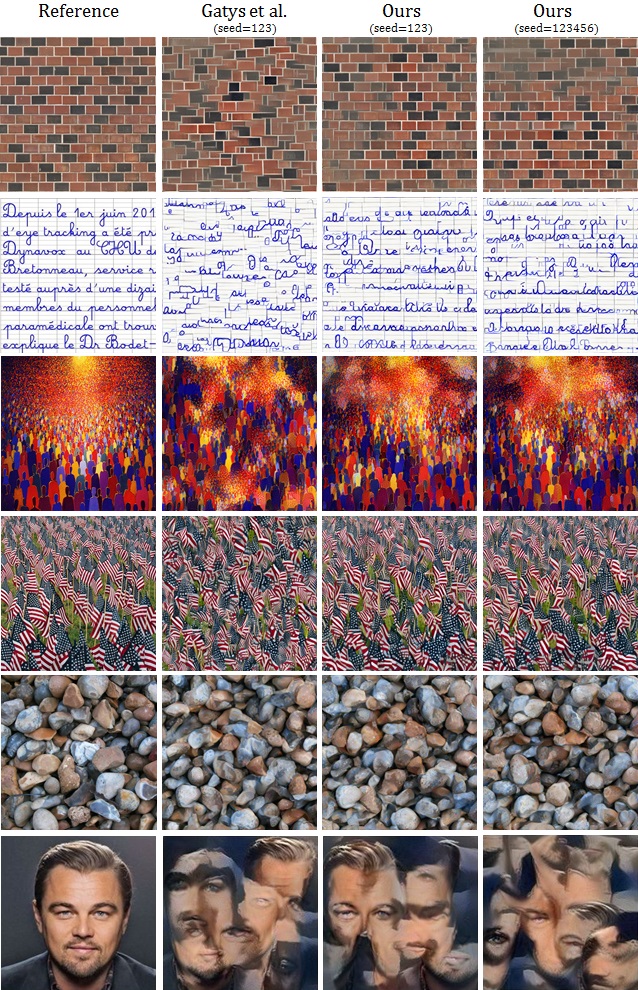}
		\end{center}
		\caption{Some results of our approach compared with \cite{Gatys2015b}. Only the initialization differs in the last two columns. Further results are shown in the supplementary material.}
		\label{demo}
	\end{figure}

	Figure~\ref{demo} shows the result of our approach applied to various structured and 
	unstructured textures. It demonstrates that the method is effective at capturing 
	long-range correlations without simply copying the content of the original texture. For instance, note how our model captures the depth aspect of the reference image in the third and fourth rows. 
	
	The problem of synthesizing near-regular structures is challenging because stochasticity and regularity are adversarial properties \citep{Lin_2006_5364}.
	Non-parametric patch-based techniques, such as \cite{Efros01}, are better suited for this task because they can \textit{tile}\footnote{Copy and paste patches side by side.} the reference image. On the other hand, regular structures are usually more problematic for parametric statistical models. Nevertheless, the two first rows of Figure~\ref{demo} demonstrate that our approach can produce good visual results and can reduce the gap to patch-based methods on these kinds of texture.
	
	Even if the reference image is not a texture, the generated images in the last 
    row (Leonardo Dicaprio's face) provide a good visual illustration of the effect of translation terms. 
	In contrast to Gatys et al., our approach preserves longer-range 
	structure, such as the alignment and similar appearance of the eyes,  
	hair on top of the forehead, the chin below the mouth, etc.  
	Finally, when the reference texture is unstructured 
	(fifth row), 
	our solution does not necessarily provide a benefit, but it also does not hurt 
	the visual quality or the diversity of the generated textures. 
	
	
	
	

	\subsection{Inpainting application}
	Modelling long-range correlations can make it possible to apply texture generation to 
	inpainting, because it allows us to impose consistency constraints between the newly 
	rendered region and the unmodified parts of the original image. 
	
	\begin{figure}[t!]
		\begin{center}
			\includegraphics[height=5.5cm]{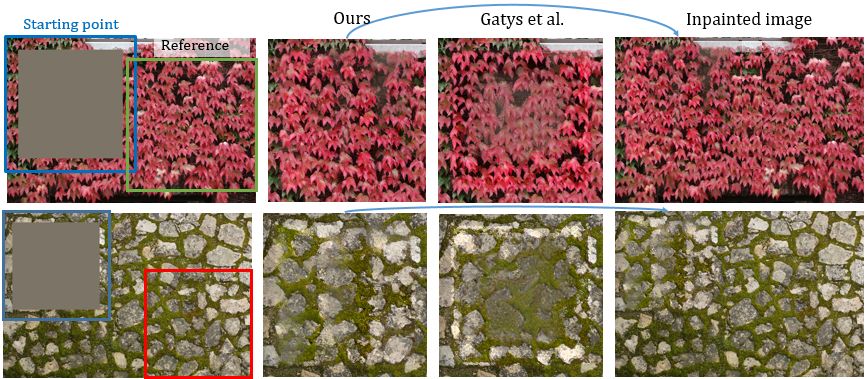}
		\end{center}
		\caption{Texture generation applied to in-painting. More in-painted images can be found in the supplementary material.}
		\label{inpainting}
	\end{figure}
	
	\begin{figure}[t!]
		\begin{center}
			\includegraphics[width=12cm]{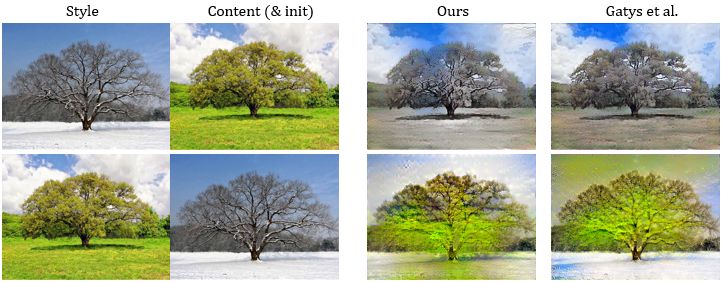}
		\end{center}
		\caption{Season transfer examples.}
		\label{season_transfer}
	\end{figure}

	To apply our approach to texture inpainting, we extracted two patches from the original 
	image: one that covers the whole area to inpaint, 
	and another one that serves as the reference texture. Then, approximately the same process as 
	for texture generation is used, with the following two modifications: 
    First, instead of random noise, the optimization starts 
	from the masked content patch (the one to inpaint) showing a grey area and its  
	non-missing surrounding. 
	Second, we encourage the borders of the output to not change much with respect to the original image using an $L_2$ penalty. We apply the penalty both 
	in the Gatys et al. rendering and in ours.
	Some inpainted images are shown in Figure~\ref{inpainting}.
	As seen in the figure, our solution significantly outperforms Gatys et al. in 
	terms of visual quality. Further results are shown in the supplementary material.
	
	\subsection{Season transfer} 
	
	Figure~\ref{season_transfer} shows the result of applying our approach to a style transfer task, as in \cite{Gatys2015c}: transferring the ``season'' of a landscape image to another one. 
	On this task, the results from our approach are similar to those from \cite{Gatys2015c}. 
	Nevertheless, in contrast to the Gatys et al. results, our approach seems to better capture global 
	information,  such as sky color and leaves (bottom row), or the appearance of branches 
	in the winter image (top row). 
	
	
	\subsection{Incorporating other types of structure}
	While we focused on feature map translations in most of our experiments, 
	other transformations can be applied as well. 
	To illustrate this point, we explored a way to generate symmetric textures using 
	another simple transformation. To this end, we propose flipping one of the two 
	feature maps before computing the Gram matrices:
	$
	G^{l}_{lr,ij} = \langle\,F^l_{i:},T_{lr}\left(F^l_{j:}\right)\rangle
	$.
	Here $T_{lr}$ corresponds to the left-right flipping operation, but we also considered up-down flipping of feature 
	maps: $\mathcal{L}^{l} = \mathcal{L}^{l}_{style} + \mathcal{L}^{l}_{lr} + \mathcal{L}^{l}_{ud}$.
	
\begin{figure}[t!]
		\begin{center}
			\includegraphics[width=11cm]{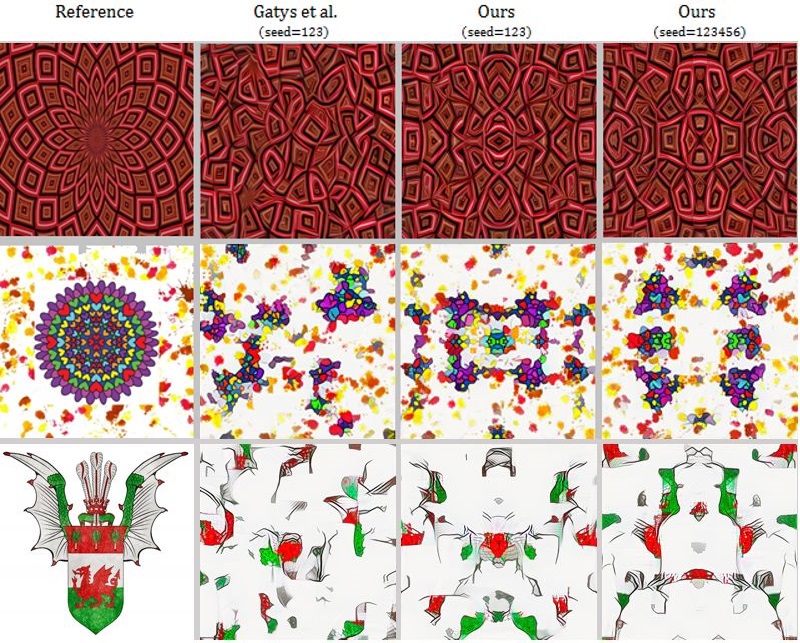}
		\end{center}
		\caption{Generation of abstract symmetric textures.}
		\label{symmetry}
	\end{figure}

	As can be seen in Figure~\ref{symmetry}, in contrast to Gatys et al., the additional loss terms capture which objects are symmetric in the reference texture, and enforce 
	these same objects to be symmetric in the reconstruction as well. 
	Other kinds of transformation could be used, depending on the type of property in the source texture one desires to preserve.
	
	\section{Conclusion}
	\label{conclusion}
	We presented an approach to satisfying long-range consistency constraints in the 
	generation of images. It is based on a variation of the method by Gatys et al., 
	and considers spatial co-occurences of local features (instead of only 
	co-occurences across features). 
	We showed that the approach permits to generate textures with various 
	global symmetry properties and that it makes it possible to apply texture generation 
	to in-painting. 
	Since it preserves correlations across sites, the approach is reminiscent of an MRF, but
    in contrast to an MRF or other graphical models, it defines correlation-constraints on 
    high-level features of a (pre-trained) CNN rather than on pixels. 
	
	{\small 
		\bibliography{biblio}
		\bibliographystyle{iclr2017_conference}	
	}
	
	\clearpage
	
	\renewcommand\appendixname{Supplementary material}
	\renewcommand\appendixpagename{Supplementary material}
	
	\appendix
	\appendixpage
	
	\section{Texture generation}
	
	\begin{figure}[h!]
		\begin{center}
			\includegraphics[width=14cm]{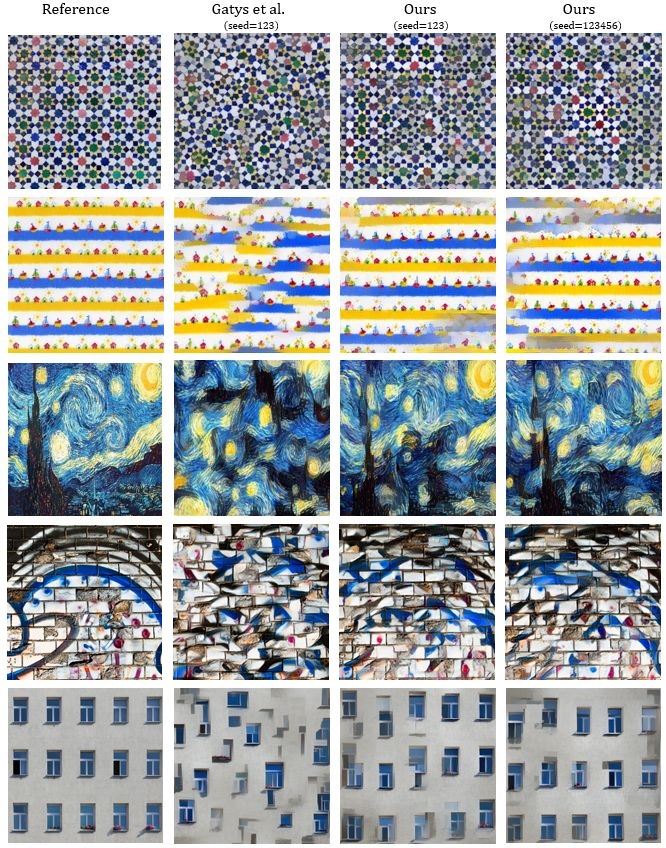}
		\end{center}
	\end{figure}
	
	\begin{figure}[h!]
		\begin{center}
			\includegraphics[width=14cm]{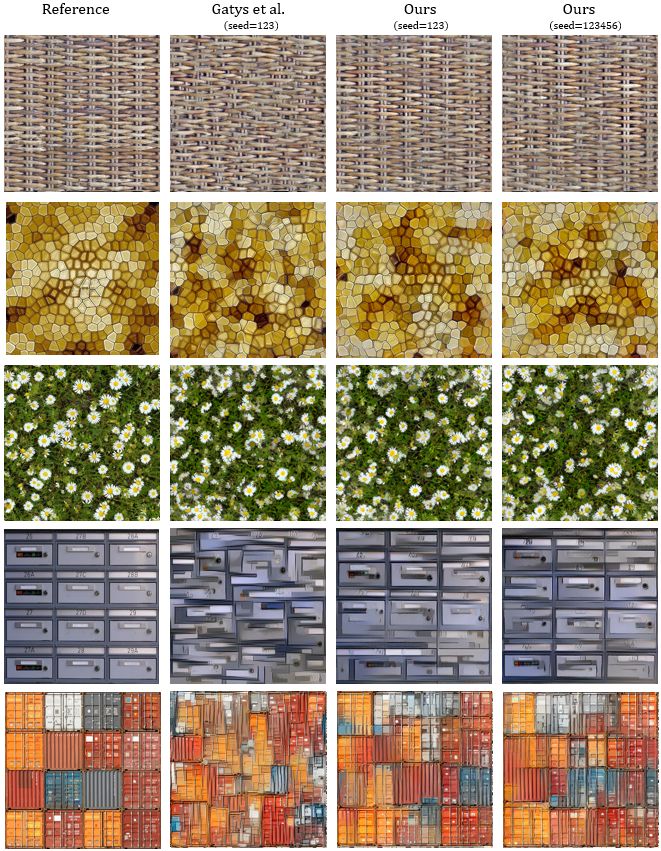}
		\end{center}
	\end{figure}
	
	\clearpage
	
	\section{Inpainting}
	
	\begin{figure}[h!]
		\begin{center}
			\includegraphics[width=14cm]{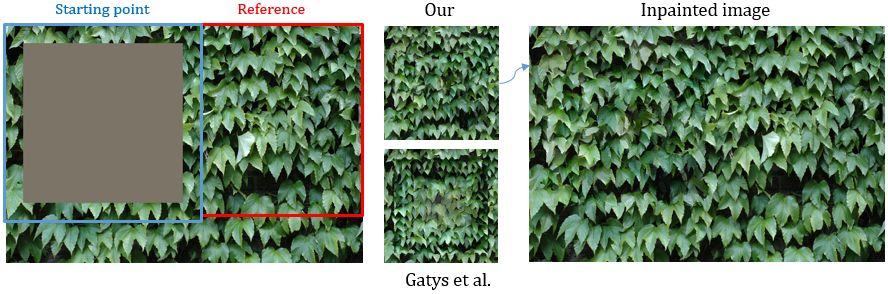}
		\end{center}
	\end{figure}
	
	\begin{figure}[h!]
		\begin{center}
			\includegraphics[width=11cm]{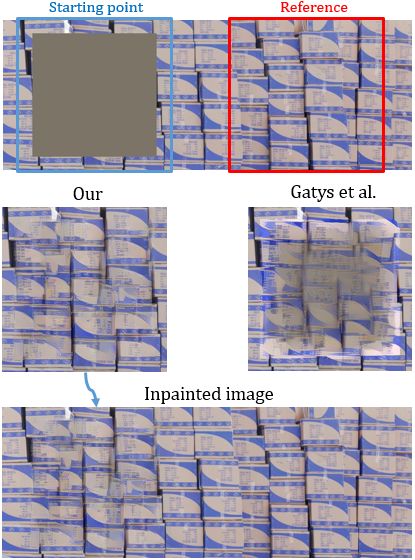}
		\end{center}
	\end{figure}
	
	\begin{figure}[h!]
		\begin{center}
			\includegraphics[width=13cm]{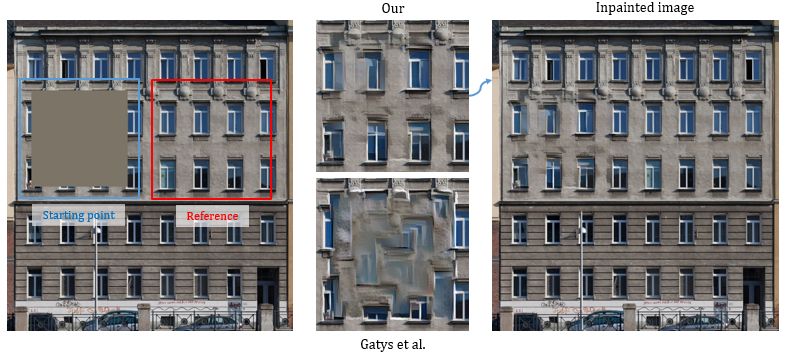}
		\end{center}
	\end{figure}
	
	\begin{figure}[h!]
		\begin{center}
			\includegraphics[width=13cm]{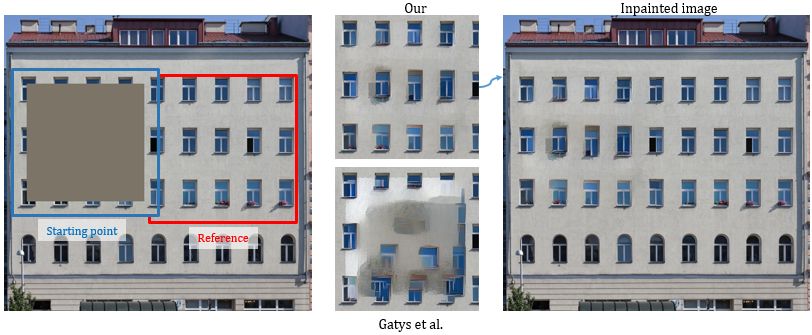}
		\end{center}
	\end{figure}
	
	\begin{figure}[h!]
		\begin{center}
			\includegraphics[width=13cm]{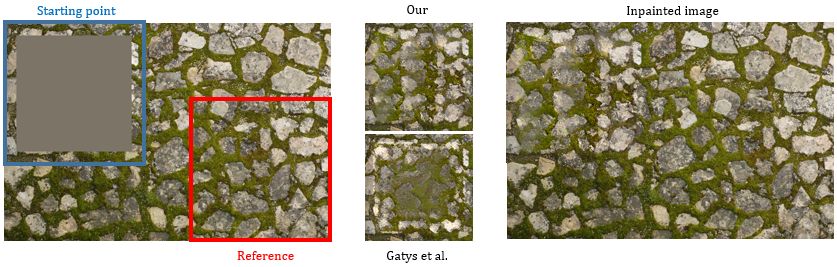}
		\end{center}
	\end{figure}
	
	\begin{figure}[h!]
		\begin{center}
			\includegraphics[width=13cm]{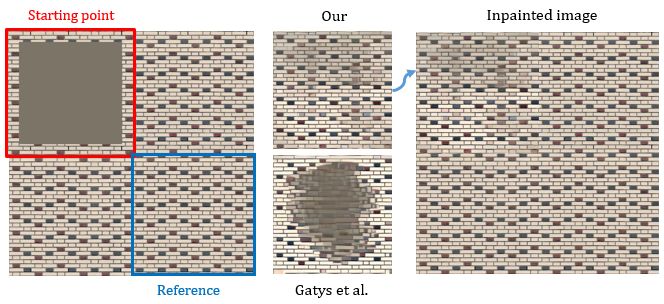}
		\end{center}
	\end{figure}
	
	\begin{figure}[h!]
		\begin{center}
			\includegraphics[width=13cm]{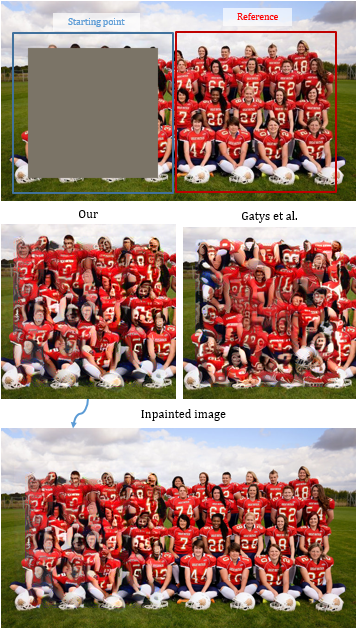}
		\end{center}
	\end{figure}
	
	\clearpage
	
	\begin{figure}[h!]
		\begin{center}
			\includegraphics[height=5cm]{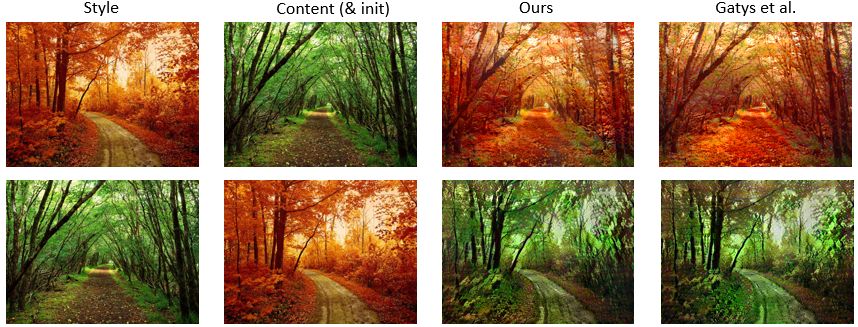}
		\end{center}
	\end{figure}
	
	\begin{figure}[h!]
		\begin{center}
			\includegraphics[width=13cm]{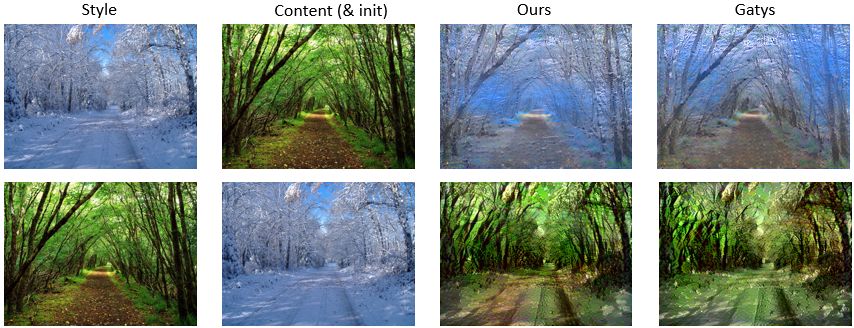}
		\end{center}
	\end{figure}
	
	\begin{figure}[h!]
		\begin{center}
			\includegraphics[width=13cm]{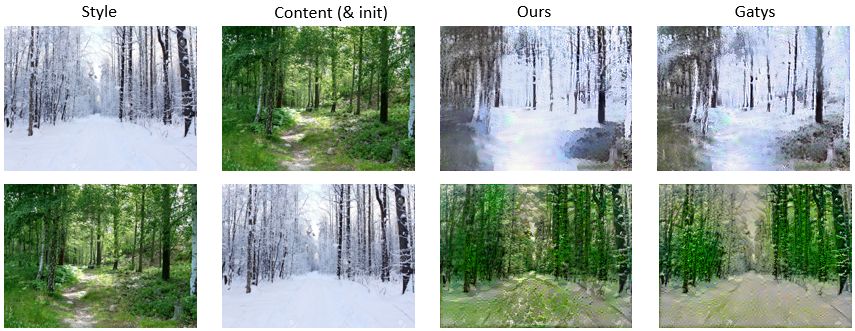}
		\end{center}
	\end{figure}
	
	\begin{figure}[h!]
		\begin{center}
			\includegraphics[width=13cm]{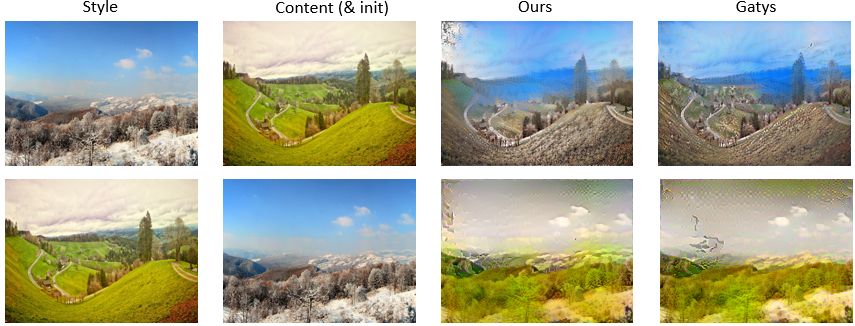}
		\end{center}
	\end{figure}
	
	\begin{figure}[h!]
		\begin{center}
			\includegraphics[width=13cm]{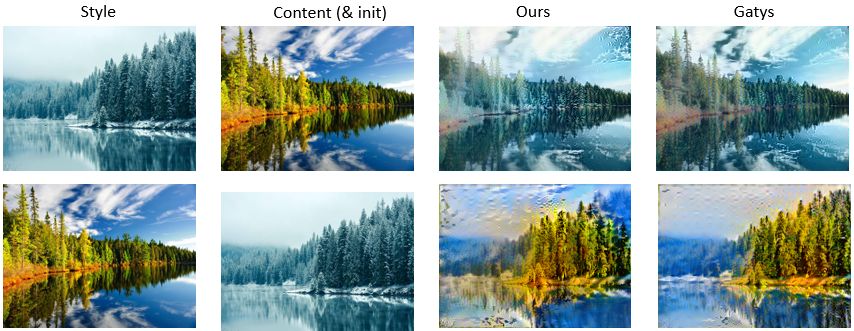}
		\end{center}
	\end{figure}
	
	\begin{figure}[h!]
		\begin{center}
			\includegraphics[width=13cm]{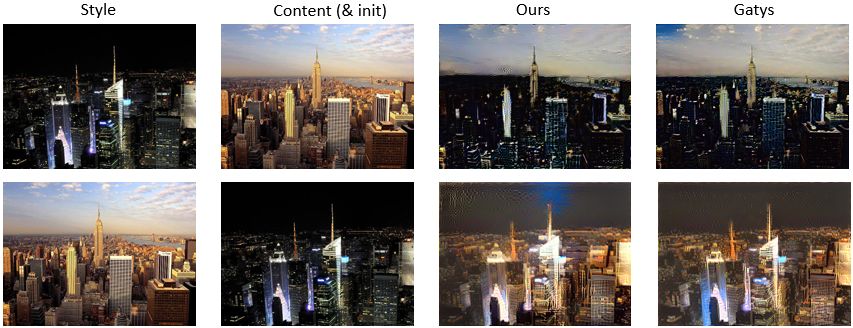}
		\end{center}
	\end{figure}
	
	\clearpage
	
	\section{Symmetric textures}
	
	\begin{figure}[h!]
		\begin{center}
			\includegraphics[width=13cm]{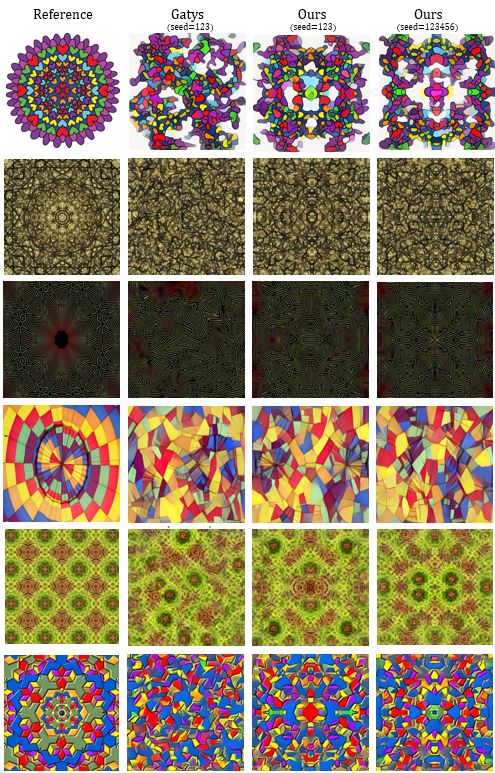}
		\end{center}
	\end{figure}
	
\end{document}